\documentclass{article} 
\usepackage{iclr2021_conference,times}

\usepackage{amsmath,amsfonts,bm}

\def\eqref#1{equation~\ref{#1}}

\def\1{\bm{1}}

\def\ra{{\textnormal{a}}}

\def\rx{{\textnormal{x}}}

\def\rva{{\mathbf{a}}}

\def\erva{{\textnormal{a}}}

\def\ervx{{\textnormal{x}}}

\def\rmA{{\mathbf{A}}}

\def\vmu{{\bm{\mu}}}
\def\vtheta{{\bm{\theta}}}
\def\va{{\bm{a}}}

\def\ve{{\bm{e}}}

\def\vx{{\bm{x}}}

\def\eva{{a}}

\def\mA{{\bm{A}}}

\def\mH{{\bm{H}}}
\def\mI{{\bm{I}}}
\def\mJ{{\bm{J}}}

\def\mX{{\bm{X}}}

\def\mSigma{{\bm{\Sigma}}}

\DeclareMathAlphabet{\mathsfit}{\encodingdefault}{\sfdefault}{m}{sl}
\SetMathAlphabet{\mathsfit}{bold}{\encodingdefault}{\sfdefault}{bx}{n}
\newcommand{\tens}[1]{\bm{\mathsfit{#1}}}
\def\tA{{\tens{A}}}

\def\tX{{\tens{X}}}

\def\gG{{\mathcal{G}}}

\def\sA{{\mathbb{A}}}
\def\sB{{\mathbb{B}}}

\def\sS{{\mathbb{S}}}

\def\emA{{A}}

\newcommand{\etens}[1]{\mathsfit{#1}}

\def\etA{{\etens{A}}}

\newcommand{\E}{\mathbb{E}}

\newcommand{\R}{\mathbb{R}}

\newcommand{\KL}{D_{\mathrm{KL}}}
\newcommand{\Var}{\mathrm{Var}}

\newcommand{\Cov}{\mathrm{Cov}}

\newcommand{\normltwo}{L^2}
\newcommand{\normlp}{L^p}

\newcommand{\parents}{Pa} 

\usepackage{hyperref}
\usepackage{url}
\usepackage{graphicx}
\usepackage[detect-all]{siunitx}
\usepackage{booktabs}
\usepackage{tablefootnote}
\usepackage{subcaption}
\usepackage{adjustbox}
\usepackage{layouts}
\usepackage{colortbl}
\definecolor{Gray}{gray}{0.9}

\newcommand{\GW}[1]{\textcolor{blue}{#1}}

\title{Towards Sustainable Census Independent Population Estimation in Mozambique}

\author{Isaac Neal$^1$, Sohan Seth$^1\ast$, Gary Watmough$^1$, Mamadou Saliou Diallo$^2$\\
University of Edinburgh$^1$ and UNICEF$^2$\\
\texttt{\{ineal,sseth,gary.watmough\}@ed.ac.uk, mamsdiallo@unicef.org}
}

\newcommand{\aMeAPE}{\textsc{aMeAPE}}
\newcommand{\MeAPE}{\textsc{MeAPE}}
\newcommand{\AggPE}{\textsc{AggPE}}
\newcommand{\MAE}{\textsc{MeAE}}

\iclrfinalcopy 
\begin{document}

\maketitle

\begin{abstract}
Reliable and frequent population estimation is key for making policies around vaccination and planning infrastructure delivery. Since censuses lack the 
spatio-temporal resolution required for these tasks, census-independent approaches, using remote sensing and microcensus data, have become popular. We estimate intercensal population count in two pilot districts in Mozambique. To encourage sustainability, we assess the feasibility of using publicly available datasets to estimate population. We also explore transfer learning with existing annotated datasets for predicting building footprints, and training with additional `dot' annotations from regions of interest to enhance these estimations. We observe that population predictions improve when using footprint area estimated with this approach versus only publicly available features.
\end{abstract}

\section{Introduction}
Accurate fine scale population estimates serve as a fundamental tool for policymakers. Many decisions involving access to services, distribution of vaccines and disaster relief, tracking of migration, and more are informed based on the most up to date population estimates for a region. Where these estimates are of insufficient resolution --- either spatially or temporally --- optimal decision making becomes difficult. 
Thus, there is a need for accurate and sustainable fine scale estimates of population globally, particularly in response to the COVID-19 pandemic, which requires efficient distribution of vaccines to vulnerable people, see e.g.,  \cite{wang_global_2020}.

Over time, census data loses its resolution, both temporal and spatial, making it difficult to inform decision making for many of the problems mentioned above. Censuses are conducted infrequently, typically decennially, and due to privacy constraints, census data is typically unavailable to the wider community at the smallest administrative levels \citep{wardrop2018spatially}. Furthermore, census data can be inaccurate or incomplete due to budget limitations, lack of training, and socio-political circumstances \citep{wardrop2018spatially}, and may quickly become outdated due to conflicts, rapid migration or urban development, and disasters  \citep{engstrom_estimating_2020}.

Census-independent population estimation (or bottom-up estimation) uses updated demographic information in periodic household surveys, or \emph{microcensuses}, and detailed visual information offered by remote sensing technology to predict intercensal population density in non-surveyed areas. This approach could improve both temporal and spatial resolution of census data,
since remote sensing data is available on a regular basis with sufficient resolution to estimate population at a fine scale.
A summary of recent works in this area is provided in Table~\ref{tab:bottomuppop}.

In this paper, we present ongoing work on bottom-up population estimation in Mozambique with a focus on \emph{sustainability}. The goal is to establish a pipeline that can be maintained and used by non-experts over a long period, that relies on existing tools and datasets as much as possible, and requires minimal human supervision, e.g., in terms of annotation. We assess the feasibility of this approach, and find that building footprint area is a crucial attribute for estimating population. Moreover, building footprint area estimates using existing deep architecture and transfer learning are improved if additional annotated data is used from the region of interest (ROI). For greater sustainability, we suggest using easier `dot' annotation \citep{lempitsky2010learning} for buildings from the ROI rather than more time-consuming polygon annotation which requires more human supervision.

\begin{figure}
    \centering
    \includegraphics[width=\linewidth]{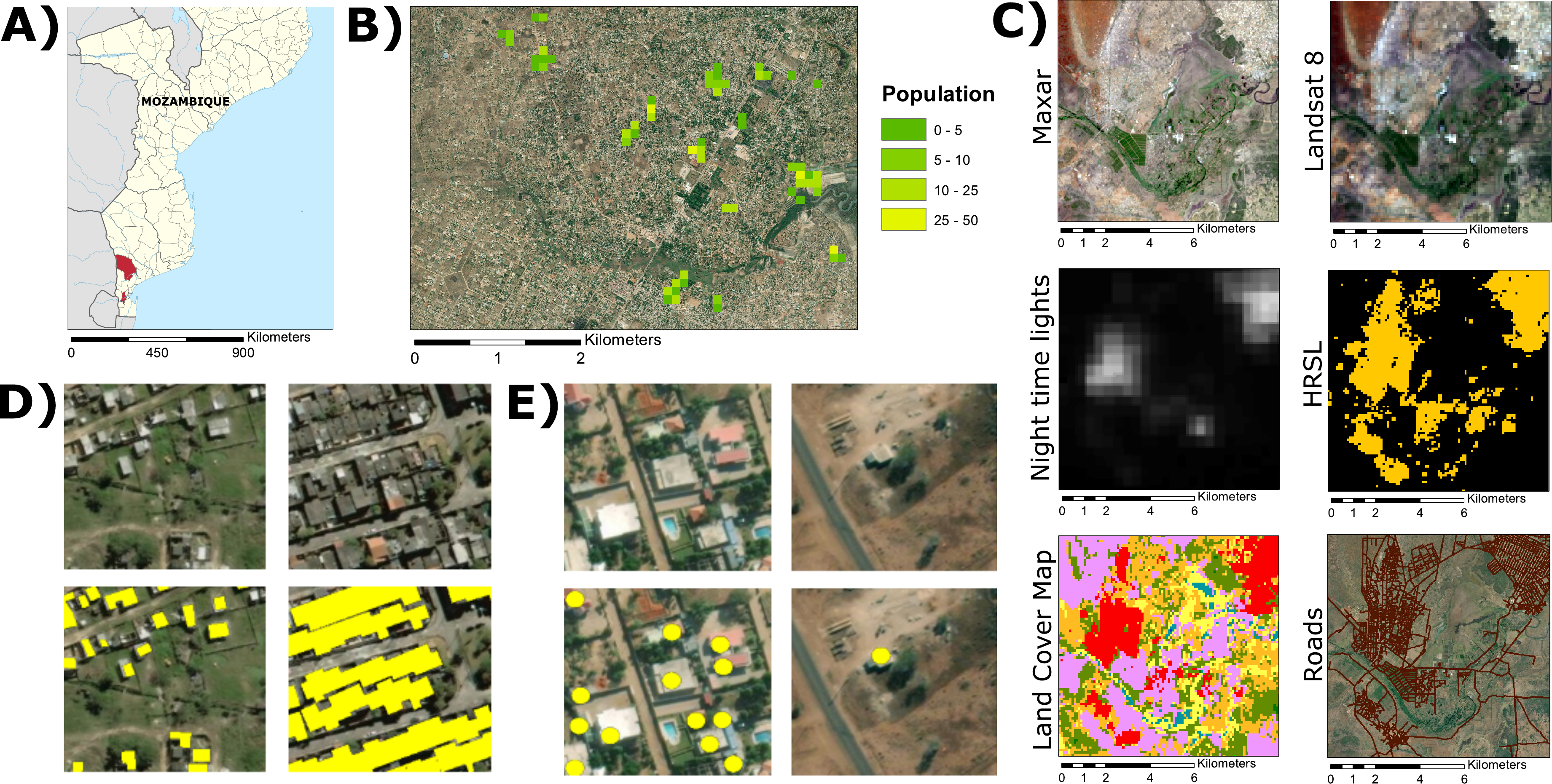}
    \caption{\textbf{A)} Regions of Mozambique (red) where microcensus was conducted, \textbf{B)} Distribution of gridded microcensus data in Boanne (BOA), \textbf{C)} Remote sensing data sources, \textbf{D)} Polygon building annotation from SpaceNet (\SI{100}{m} tiles), \textbf{E)} Dot building annotation from Mozambique (\SI{100}{m} tiles).}
    \label{fig:papersummary}
\end{figure}

\section{Data and Processing}
\textbf{Microcensus:}
UNICEF funded a microcensus in 2019 conducted by SpaceSUR and GroundWork over two ROIs in Mozambique: Boanne (BOA) and Magude (MGD) (see Fig. \ref{fig:papersummary}A). The survey was conducted at a household level and households were exhaustively sampled over several primary sampling units (PSUs). We aggregated the household survey data to a \SI{100}{m} grid to generate the response variable for predicting \emph{gridded} population from remote sensing data (see Fig. \ref{fig:papersummary}B).

\textbf{Remote Sensing:}
A variety of remote sensing imagery and derived data were used to model population count. The data sources are summarized in Table \ref{tab:datasources}.  Example images of each data source are shown in Fig. \ref{fig:papersummary}C. The \SI{50}{cm} high-resolution satellite imagery (Vivid from Maxar) was a mosaic of WorldView-2 images, mostly from 2018 and 2019 (83\% and 17\% for BOA and 43\% and 33\% for MGD, remainder from 2011 to 2020). The night time lights (NTL) data was average intensity for the year 2019. The land cover classification was reclassified to condense forest types to \textit{open forest} and \textit{closed forest}. We ignored \textit{no class}, \textit{closed forest}, \textit{herbaceous wasteland}, \textit{moss}, \textit{bare}, \textit{snow}, \textit{water}, and \textit{sea} as they did not appear in our survey tiles. This left 5 classes. All raster data was reprojected to the \SI{100}{m} survey grid for each ROI. 
Three types of resampling were used for different data-sources: average resampling for continuous data, nearest-neighbor resampling for coarser resolution categorical data, and any resampling (i.e. 1 if any underlying tile is 1 and otherwise 0) for finer resolution binary data.

\begin{table}[t]
\centering
\caption{Summary of recent literature on topic of bottom up population estimation}.
\label{tab:bottomuppop}
\resizebox{\linewidth}{!}{
    \begin{tabular}{@{}lllll@{}}
        \toprule
        & \citet{weber2018census} & \citet{engstrom_estimating_2020} & \citet{hillson_estimating_2019} & \citet{leasure_national_2020} \\
        \cmidrule(l){2-5}
        \rowcolor{Gray}
        Region of Interest & Nigeria & Sri Lanka & Bo, Sierra Leone & Nigeria \\ 
        Input Resolution & \SI{0.5}{m} (Maxar) & \begin{tabular}[c]{@{}l@{}}\SI{10}{m} (Polygon data)\\ 12-\SI{30}{m} (Urban Footprint)\\
        \SI{750}{m} (Night time Lights) \end{tabular} & \SI{30}{m} (Landsat) & \begin{tabular}[c]{@{}l@{}}0.5m (Maxar), \SI{100}{m}  \\(WorldPop), various \\(OSM school density, \\ household size)\end{tabular} \\ \addlinespace
        \rowcolor{Gray}
        Output Resolution & \SI{90}{m} & Village level & City district level & \SI{100}{m} \\ \addlinespace
        Input Data Cost & High (Maxar data) & \begin{tabular}[c]{@{}l@{}}Free (public data)\\ High (Maxar data)\end{tabular} & Free (public data) & \begin{tabular}[c]{@{}l@{}}Free (OSM, WorldPop)\\ High (Maxar data)\end{tabular} \\ \addlinespace
        \underline{Performance} \\ \addlinespace
        \rowcolor{Gray}
        Validation & eTally data & Train/test split & LOOCV & Train/test split  \\ 
        \MeAPE & - & 28 & 11 & - \\ 
        \rowcolor{Gray}
        $R^2$ & 0.98 & 0.58 & - & 0.26 \\ \addlinespace
        \bottomrule
    \end{tabular}
}
\captionsetup{width=0.8\linewidth,font=scriptsize}
\caption*{NOTE: LOOCV = leave one out cross validation, \MeAPE = median absolute percent error}
\vspace{-1.5em}
\end{table}

\begin{table}[t]
\centering
\caption{Remote sensing data sources, their characteristics and features extracted.}
\label{tab:datasources}
\resizebox{\linewidth}{!}{
\begin{tabular}{lcccccc}
\toprule
    Data & Resolution & Year & Frequency & Prerequisite & \begin{tabular}{@{}c@{}}Publicly avail. \\ (source) \end{tabular} & \begin{tabular}{@{}c@{}}Features\\Extracted (Count) \end{tabular} \\
\midrule
    \rowcolor{Gray}
    Vivid Imagery & \SI{0.5}{m} & Various & \SI{4}{\day} & N/A & No (Maxar) & Building area (1) \\ 
    Landsat 8 & \SI{30}{m} & 2019 & \SI{8}{\day} & N/A & \begin{tabular}[c]{@{}c@{}} Yes (NASA/\\USGS)\end{tabular} & \begin{tabular}[c]{@{}c@{}}10 bands, NDVI,\\ NDWI (12)\end{tabular} \\ 
    \rowcolor{Gray}
    \begin{tabular}[c]{@{}l@{}}High Resolution\\ Settlement Layer\end{tabular} & \SI{30}{m} & 2015 & N/A & \begin{tabular}[c]{@{}c@{}}High res.\\ imagery and\\ census data\end{tabular} & \begin{tabular}[c]{@{}c@{}}Yes\\(CIESIN)\end{tabular} & Binary Map (1) \\ 
    \begin{tabular}[c]{@{}l@{}}Land cover \\ classification\end{tabular} & \SI{100}{m} & 2019 & Annual & Various & Yes (ESA) & LCC (5) \\
    \rowcolor{Gray}
    Night-time lights & \SI{750}{m} & 2019 & \SI{1}{\day} & N/A & \begin{tabular}[c]{@{}c@{}}Yes\\(NOAA)\end{tabular} & Radiance (1) \\ 
    Road Data & Vector & Various & {N/A} & \begin{tabular}[c]{@{}c@{}}Volunteer \\ annotation\end{tabular} & Yes (OSM) & \begin{tabular}[c]{@{}c@{}}Distance to \\ road (1)\end{tabular}\\
\bottomrule
\end{tabular}
}
\captionsetup{width=0.98\linewidth,font=scriptsize}
\caption*{NOTE: Although WorldView-2 imagery may be available every 4 days, we only had access to Vivid data from Maxar collected from various years. The number in parenthesis in the final column shows the number of derived features from that dataset.}
\vspace{-2em}
\end{table}
\textbf{Distance to Road Calculation:}
We rasterized the OSM road shapefiles to produce a Boolean raster indicating the presence of road in each grid tile, and calculated the Euclidean distance from each tile to the nearest tile with road as distance to road. We treated all road annotations from the OSM dataset the same, yielding one value for distance to road for each grid tile.

\textbf{Building Footprint Identification}
We used a U-Net architecture \citep{ronneberger_u-net_2015} with a VGG16 encoder \citep{simonyan_very_2015} pre-trained on weights from ImageNet \citep{imagenet} to segment an image into \textit{building} and \textit{non-building} areas. \textbf{BFIa)} We pretrained the building footprint segmentation model on annotated satellite imagery from the SpaceNet challenge \citep{van_etten_spacenet_2019}. We used the processed AOI 1 dataset from SpaceNet v1, which contains a large quantity of satellite imagery from Brazil with polygon building annotations. The dataset differs from our ROIs in Mozambique in several ways: SpaceNet tiles have more urban areas, a wetter climate, and different building colouration (see Figs.~\ref{fig:papersummary}D-E). These factors hindered transfer learning between the tasks, and justified fine-tuning on a dataset from Mozambique. \textbf{BFIb)} We fine-tuned the model on `dot' annotated buildings covering an area of \SI{10}{km^2} (i.e., 1,000 grid tiles to reduce manual annotation) from the ROIs (outside surveyed areas to avoid data leakage). These annotations are a subset of those produced through a mixture of automatic estimation and manual curation in previous work by SpaceSUR. The `dot' annotation was converted to raster class labels by rasterizing a circle around each point with the average area of buildings in the respective ROI. Dot annotation was preferred over polygon annotations to avail sustainability since the former is less time-intensive than the latter. This approach might not infer the exact boundary of a building but is only an intermediate step to population estimation.

\textbf{Context Variable}
We introduced the notion of \emph{context} to our model by calculating the average of each extracted feature (except distance to road) in the tiles surrounding the tile of interest (two surrounding contexts, 8 tiles and 24 tiles), and using these values as additional features in the model. In doing so, we aimed to provide the model with an understanding of the features on coarser scales.

\textbf{Non-representative Tiles}
Upon investigating the data, we concluded that, due to the retrospective nature of the data, several tiles at the boundary of the PSUs only covered a PSU partially, and several surveyed areas had developed either before or after our high resolution satellite imagery was captured. This meant that the tiles often contained non-surveyed buildings or missing buildings, and therefore, our gridded data contained both developed tiles (i.e., with a large number of buildings) labeled as low population, and undeveloped tiles labeled as high population. Although it is possible to either automatically detect these `outlier' tiles during training or use a robust loss function that is insensitive to them, it is not possible to validate a model on non-representative data. Therefore, we manually excluded these grid tiles from the data by comparing the surveyed buildings (GPS locations available in the microcensus) with those appearing in the Vivid imagery.

\section{Results and Discussion}
\textbf{Dataset} We have 199 survey grid tiles and 61 predictor variables (see Table~\ref{tab:datasources}).
\textbf{Cross-validation}
Due to the limited quantity of data, we evaluated our population model through 4-fold cross validation. The data is split spatially into four approximately equal sized subsets (for each ROI separately), and we reported the error metrics over \emph{pooled} prediction from the four validation folds.
\textbf{Model} Due to the limited availability of data, we fit a single linear model with Huber loss function and $\ell_1$ regularization \citep{yi_semismooth_2017} for both districts to model the log of population count given a combination of public features and BFIa or BFIb. We chose the model hyperparameters using 3-fold cross-validation over the training set for each fold of our spatial cross validation. \textbf{Evaluation Metrics} We chose several evaluation metrics, i.e., $ R^2 = 1- {\sum_i (y_i - \hat{y})^2}/{\sum_i (y_i - \bar{y})^2}$, 
median absolute error $\MAE = \textrm{median}\, |y_i - \hat{y}_i|$, 
median absolute percentage error $\MeAPE = \textrm{median}\, {|y_i - \hat{y}_i|}/{y_i}$, 
adjusted median absolute percentage error $\aMeAPE = \textrm{median}\, {|y_i - \hat{y}_i|}/{(y_i + 10)}$ (to avoid division by zero), and 
aggregated percentage error $\AggPE(A) = {|\sum_{i \in A} y_i - \sum_{i \in A} \hat{y}_i|}/{\sum_{i \in A} y_i}$ (to capture error at a ROI level).
\textbf{Null model} The null model predicted the population as the mean of the training set irrespective of the feature values.
\textbf{Results} We observe that the model can effectively predict population, and outperforms the null model. The model performs the best with either public and fine-tuned building footprints (BFIb) as features, or only BFIb as features, and the performances are similar. A loss in accuracy is incurred when using either public only or public and pre-trained building footprints (BFIa) as features, however, the accuracy is still acceptable ({\MAE} is 3.8 versus 7.5 in null model). The model performed the worst when using only pre-trained building footprints (BFIa) as features, indicating that fine-tuning improves performance substantially.

\begin{minipage}{0.6\linewidth}
    \centering
    \captionof{table}{Summary of model performance}
        \resizebox{\linewidth}{!}{
            \begin{tabular}{lrrrrr}
                \toprule
                Features used & \multicolumn{1}{c}{$R^2$} & \multicolumn{1}{c}{$\MeAPE$} & \multicolumn{1}{c}{$\aMeAPE$} & \multicolumn{1}{c}{$\MAE$} & \multicolumn{1}{c}{$\AggPE$} \\ 
                \cmidrule(l){1-1} \cmidrule(l){2-6}
                \rowcolor{Gray}
                Public          &   0.05 &  51.8\% &  0.23 &  3.84 &  25.4\% \\
                BFIa            &  -0.08 &  59.9\% &  0.25 &  4.02 &  32.1\% \\
                \rowcolor{Gray}
                BFIb            &   0.54 &  39.2\% &  0.20 &  3.41 &  14.9\% \\
                Public + BFIa   &   0.05 &  50.1\% &  0.23 &  3.97 &  27.3\% \\
                \rowcolor{Gray}
                Public + BFIb   &   0.53 &  42.1\% &  0.19 &  3.45 &  13.2\% \\
                \addlinespace
                \addlinespace
                \underline{Null Model} & -0.12 & 76.45\% & 0.41 & 7.57 & 1.68\% \\
                \addlinespace
                \bottomrule
            \end{tabular}
        }
        \captionsetup{width=0.9\linewidth,font=scriptsize}
        \caption*{See \textbf{Evaluation Metrics} above for metric definitions. BFIa and BFIb are pre-trained and fine-tuned building area estimates respectively. Predicted vs. observed plot (\textbf{right}) summarizes the results for Public + BFIb.}
\end{minipage}
\hspace{0.5em}
\begin{minipage}{0.3\linewidth}
    \centering
    \includegraphics[height=0.175\paperheight]{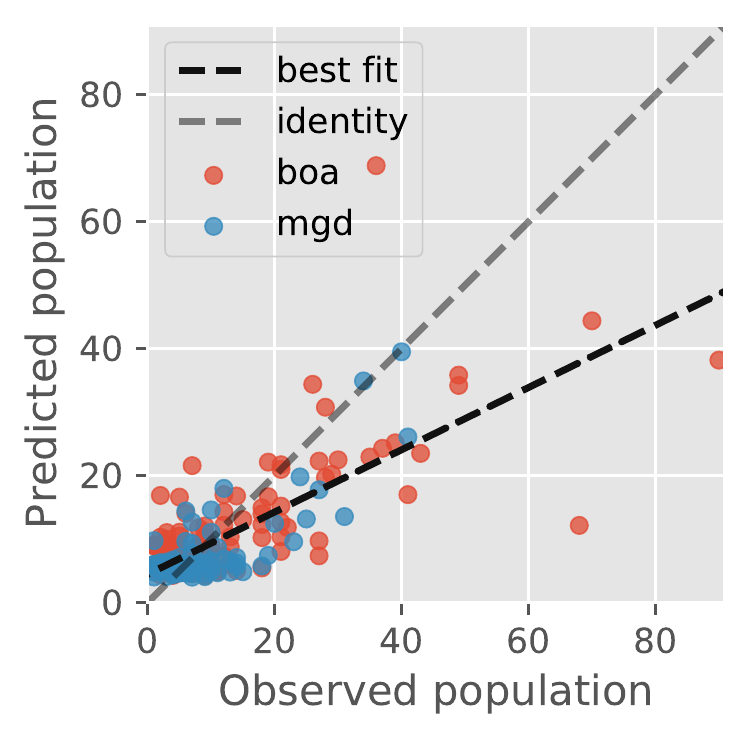}
\end{minipage}

We are currently addressing the following improvements to the model:
\textbf{Buildings under construction} Our building segmentation model often identifies buildings that are under construction or buildings that are in ruin (e.g., historical buildings from colonial era with roof caved in). Since building footprint is an important feature in predicting population this can introduce bias in the outcome. We are improving the building segmentation model to help reduce this bias. \textbf{Non-residential buildings} Our building segmentation model does not discriminate between non-residential buildings (e.g., grain storage or schools) and residential buildings. We are addressing this by computing individual footprints prior to total area and discounting buildings that are too large or too small to likely be inhabited.
\textbf{Distance to road} We treat all road types to be the same. We are extending this approach to give a more informative set of distance variables (e.g. distance to major highway) for different road types. We also intend to calculate road density to compare usefulness.
\textbf{Sustainability} Sustainability is an issue in most bottom-up estimation techniques in the literature. The causes for this are twofold: the high cost of conducting population surveys, and limitations inherent to the machine learning methodology that might require expert annotation. Our results show that population can be estimated with limited survey data and `dot' annotations. We are addressing this further by exploring ways to reduce dot annotation required.
\textbf{External validation and transferability} It is difficult to externally validate the existing population estimation approaches or to compare them. This is because existing methods have been applied on different countries and a standard microcensus dataset does not exist for comparison. Furthermore, when microcensus data does exist it is difficult to access satellite images with the same spatial, temporal and radiometric characteristics using the Vivid data product. We are addressing this through conducting fresh microcensus surveys in Mozambique.
\subsubsection*{Acknowledgments}
The project was funded by the Data for Children Collaborative with UNICEF. The authors thank Pablo Martín Amieva (SpaceSUR) for his valuable insights on the project.
\bibliography{population,buildings}
\bibliographystyle{iclr2021_conference}

\end{document}